\documentclass{article}
\usepackage{spconf,amsmath,graphicx,hyperref}
\usepackage{enumitem}
\usepackage{booktabs,multirow,array,makecell,pifont,adjustbox}
\newcommand{\cmark}{\ding{51}}
\newcommand{\xmark}{\ding{55}}
\usepackage{stfloats}

\usepackage{booktabs,multirow,tabularx,caption}
\usepackage{pifont} % 若你没有定义 \cmark / \xmark

\usepackage{amsmath,epstopdf,amssymb,color,url,multirow,multicol,verbatim,threeparttable,diagbox,makecell,booktabs,color,colortbl}  
\definecolor{mygray}{rgb}{0.9,0.9,0.9}

\title{DiffFace-Edit: A Diffusion-Based Facial Dataset for
Forgery-Semantic Driven Deepfake Detection Analysis}
\name{Feng Ding$^{1}$,
      Wenhui Yi$^{1}$,
      Xinan He$^{1}$,
      Mengyao Xiao$^{1 \star}$,
      Jianfeng Xu$^{1 \star}$,
      Jianqiang Du$^{1}$\thanks{$^{\star}$Corresponding author.\\ \hspace*{2em}This work was supported in part by the National Natural Science Foundation of China under Grant 62262041, and in part by the Jiangxi Provincial Natural Science Foundation under Grant 20232BAB202011.}}
\address{$^{1}$ NanChang University \\}

\begin{document}
%\ninept
%
\maketitle
\begin{abstract}

Generative models now produce imperceptible, fine-grained manipulated faces, posing significant privacy risks. However, existing AI-generated face datasets generally lack focus on samples with fine-grained regional manipulations. Furthermore, no researchers have yet studied the real impact of splice attacks, which occur between real and manipulated samples, on detectors. We refer to these as detector-evasive samples. Based on this, we introduce the DiffFace-Edit dataset, which has the following advantages: 1) It contains over two million AI-generated fake images. 2) It features edits across eight facial regions (e.g., eyes, nose) and includes a richer variety of editing combinations, such as single-region and multi-region edits. Additionally, we specifically analyze the impact of detector-evasive samples on detection models. We conduct a comprehensive analysis of the dataset and propose a cross-domain evaluation that combines IMDL methods. Dataset will be available at https://github.com/ywh1093/DiffFace-Edit.
% The extraordinary ability of diffusion models to generate photorealistic faces has increased disinformation risks and created demand for detectors that distinguish AI-generated fake faces from real ones. However, progress remains limited by the lack of large, modern corpora targeting fine-grained partial edits. In this paper, we introduce the DiffFace-Edit dataset, which has the following advantages: 1) Ample scale, with over two million AI-generated fake images. 2) Fine-grained content, with edits across eight facial regions (e.g., eyes, nose). 3) Detector-evasive samples, with the first systematic analysis of this sample type. We conduct a comprehensive analysis of the dataset and propose a cross-domain evaluation that combines IMDL methods. With the DiffFace-Edit dataset, researchers can effectively
% expedite the research on robust, artifact-aware facial forgery detection.
\end{abstract}

\vspace{-2mm}
\section{Introduction}
\label{sec:intro}
\vspace{-2mm}

The proliferation of AI-generated content has brought many benefits but also significant challenges, particularly in the field of digital forensics \cite{ding2025fairadapter,pang2024heterogeneous, zhou2025breaking}. Diffusion models (DMs) have become a major concern, as they can produce highly realistic synthetic faces and enable fine-grained facial manipulations. These capabilities raise serious security and privacy issues \cite{lin2024preserving,pang2025unified}.

While some recent work has focused on creating datasets of DM-generated images to aid detection, existing resources often fall short for the specific needs of facial forensics \cite{ding2021anti}, as shown in Tab.~\ref{tab:datasets-2col}. Many large-scale datasets, such as CIFake and GenImage \cite{bird2024cifake, wang2023dire, zhu2023genimage}, primarily feature everyday scenes rather than faces. Although some face-centric datasets exist \cite{cheng2024diffusion, bhattacharyya2024diffusion}, they are generally small, collected under limited conditions with little prompt diversity, and lack detailed annotations. This scarcity makes it difficult to train robust detectors that can generalize effectively to real-world facial forgeries.

To address this gap, we introduce DiffFace-Edit, a new dataset focused on diffusion-based partial facial editing. Our dataset is unique in three key ways: 1) DiffFace-Edit is the first dataset dedicated exclusively to partial facial edits generated by DMs. It covers eight distinct canonical regions (e.g., eyes, nose) and includes over two million forged face images. 2) We include detector-evasive samples \cite{ding2024disrupting, fan2025generating} that pose a significant challenge to existing IMDL localizers. To our knowledge, this is the first dataset to provide a comprehensive analysis of this challenging sample type. 3) Every forged image in DiffFace-Edit is precisely annotated with the edited region(s) and the specific prompt used for the manipulation. By providing a large, well-annotated collection of partial facial edits, including challenging, detector-evasive samples—DiffFace-Edit fills a critical void in the field. It is designed to facilitate the development of more robust and generalizable forgery detection and localization methods \cite{he2025vlforgery}.

% While current research has focused on DM-based synthetic images, state-of-the-art forensic detectors (e.g., CLIP-based methods\cite{cozzolino2024raising}) report near-perfect accuracy (99\%) on full synthesis benchmarks such as GenImage\cite{zhu2023genimage} and DRCT-2M\cite{chen2024drct}. However, recent studies indicate that this performance may be misleading. DRCT\cite{chen2024drct} discovered that detectors achieve only around 50\% accuracy on semantically similar samples (e.g., (SDv1.5-Ctrl\cite{zhang2023adding})-generated images), despite near-perfect accuracy on fully synthetic data. AIDE\cite{yan2024sanity} further demonstrated that, under real-world conditions, most advanced detectors struggle, with accuracy dropping below 10\% on the Chameleon dataset\cite{yan2024sanity}.

%This performance gap raises fundamental questions: Do current IMDL locators like PSCC-Net\cite{liu2022pscc} truly identify imperceptible forgery traces, or are they over-relying on visible content semantics? Their success may stem from: i) a focus on semantic features rather than low-level artifacts; ii) training-test distribution similarity rather than a deep understanding of forgeries.

%To systematically explore this issue, we introduce the DM-based partially synthetic face dataset (DiffFace-Edit). It is designed to drive research into detecting forgery clues that may be embedded in the content semantics themselves. 

The main contributions of this paper are summarized as follows:
\begin{itemize}[noitemsep, topsep=0pt, parsep=0pt, partopsep=0pt, leftmargin=*]
    \item A million-scale partially synthetic face dataset, featuring 6 generative models, 2 manipulation scenarios (single-region and multi-region), 3 manipulation types (splice, remove, and copy-move), and 2 sample categories (IMDLBenCo\cite{ma2024imdl}-compliant samples and semantically ambiguous samples).
    \item A cross-domain evaluation that combines IMDL methods, integrating 8 IMDL locators.
    \item We generate detector-evasive samples and, to the best of our knowledge, provide the first systematic analysis of this sample type.
\end{itemize}

\captionsetup{font=footnotesize, skip=3pt} % 标题小一点、间距更紧
\begin{table*}[!t]
\centering
\caption{Quantitative comparison of DiffFace-Edit with existing datasets.}
\label{tab:datasets-2col}

% ——整体风格：更宽的列间距与更舒适的行距——
\setlength{\tabcolsep}{6pt}    % 列间距（默认 ~6pt；双栏建议略大于 3pt）
\renewcommand{\arraystretch}{1.12} % 行距略放大，提升可读性
\footnotesize % 表内字号统一为 footnotesize（比 scriptsize 易读）

\begin{adjustbox}{width=0.9\linewidth}
    \begin{tabular}{ccccccc}
    \toprule
    % \multirow{2}{*}{Dataset} &
    % \multirow{2}{*}{Year} &
    % \multirow{2}{*}{Content} &
    % \multicolumn{2}{c}{Face Images/Videos} &
    % \multirow{2}{*}{Fine-grained Partial Edit} \\
    % \cmidrule(lr){4-5}
    %  &  &  & \#Real & \#Fake & \\
    Dataset &
    Year &
    Content &
    \#Real & \#Fake &
    Fine-grained Partial Edit &
    Detector-evasive Samples \\
    \midrule
    CIFAKE\cite{bird2024cifake} & 2023 & General & 60k   & 60k    & \xmark & \xmark \\
    DiffusionForensics-LSUN \cite{wang2023dire}   & 2023 & Bedroom & 42k   & 215k   & \xmark & \xmark \\
    DiffusionForensics-General\cite{wang2023dire} & 2023 & General & 50k   & 60k    & \xmark & \xmark \\
    GenImage\cite{zhu2023genimage}                & 2023 & General & 1331k & 1350k  & \xmark & \xmark \\
    ForgeryNet\cite{he2021forgerynet}             & 2021 & Face    & 1438k & 1458k  & \cmark & \xmark \\
    DiffusionDeepfake\cite{bhattacharyya2024diffusion} & 2024 & Face & 94k   & 100k   & \xmark & \xmark \\
    DiFF\cite{cheng2024diffusion}                 & 2024 & Face    & 23k   & 500k   & \cmark & \xmark \\
    \midrule
    \textbf{DiffFace-Edit} & \textbf{2025} & \textbf{Face} & \textbf{30k} & \textbf{2M+} & \textbf{\cmark} & \textbf{\cmark} \\
    \bottomrule
    \end{tabular}
\end{adjustbox}
\vspace{-10pt}
\end{table*}

\vspace{-2mm}
\section{DiffFace-Edit dataset}
\label{sec:DiffFace-Edit dataset}
\vspace{-2mm}

\begin{figure}[htb]
  \centering
  \includegraphics[width=0.98\linewidth]{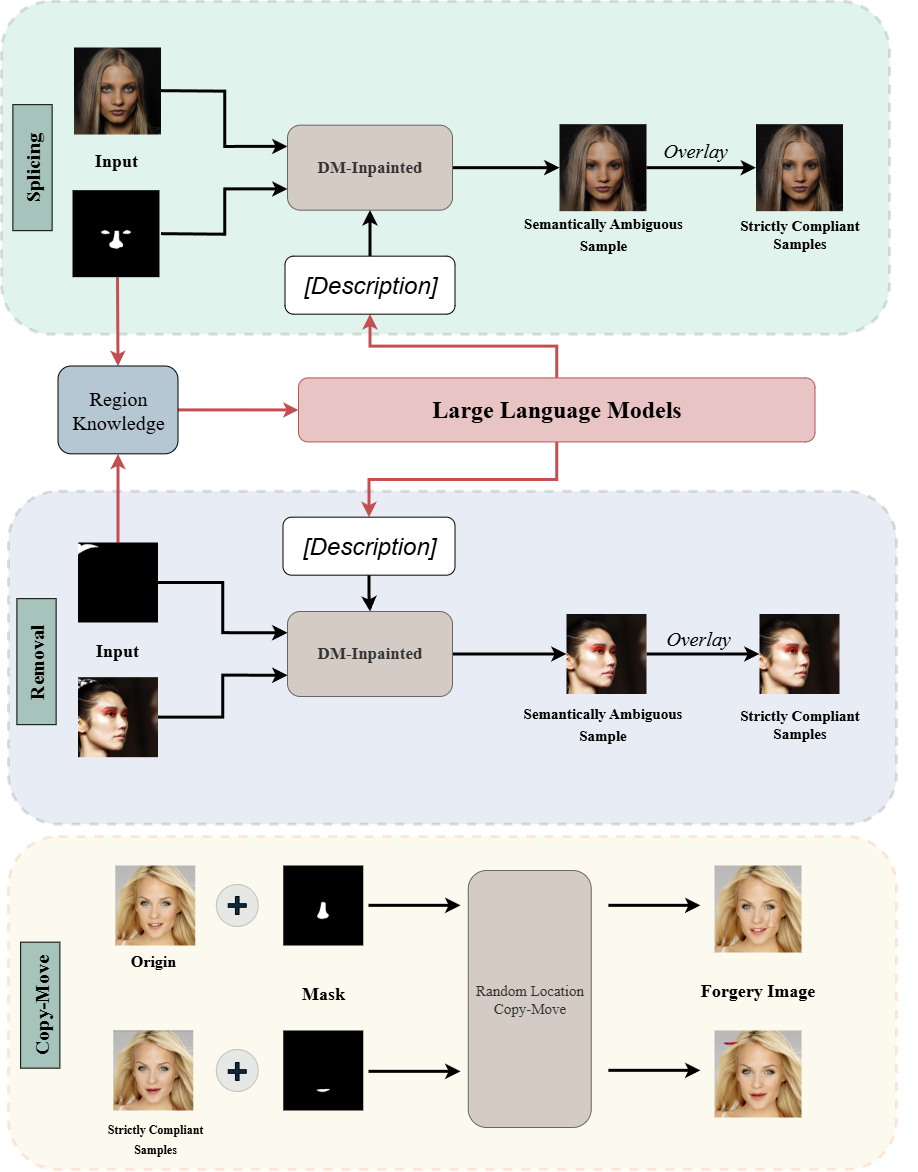}
  \vspace{2pt}
  \caption{The fake data generation pipeline of proposed DiffFace-Edit dataset.}
  \label{fig:datagen}
  \vspace{-15pt}
\end{figure}

This section outlines the process for constructing the DiffFace-Edit dataset of partially synthetic faces (see Fig.~\ref{fig:datagen}), along with its statistics.

\vspace{-1mm}
\subsection{Pristine Data Collection and Partially Synthetic Face Generation}
\label{ssec:subhead}
\vspace{-1mm}

\textbf{Real data.} We utilized the well-established facial dataset CelebAMask-HQ\cite{lee2020maskgan} as our data source, as it provides detailed annotations of various facial landmarks, including nose, eyebrows, hair, ears, eyes, and other attributes. We strictly adhered to the dataset's licensing agreement to guarantee compliance for inclusion in our dataset, as well as for secondary usage in training and testing. 

\noindent
\textbf{Partially Synthetic Face.} To ensure diversity in partially synthetic faces, we employed six distinct open-source inpainting models. We focused on generating synthetic content for eight facial landmarks—nose, eyebrows, hair, eyes, lips, mouth, neck, and ears—as these attributes consistently yield realistic results. Other attributes (e.g., skin) often caused undesirable global facial distortions despite carefully crafted prompts, and were thus excluded during generation while preserving their original appearance. Following traditional IMDL taxonomy, we categorized manipulations into three types: \textbf{1) Splicing.} Two editing scenarios were designed, single-region editing (generating one fake image per associated region for each real image) and multi-region editing (producing five fake images per real image, each containing random 2–4 manipulated regions). The `Llama-3.2-3B-Instruct' model was utilized to generate corresponding prompt templates. \textbf{2) Removal.} The essence lies in naturally replacing target regions with context-aware backgrounds. However, due to limitations of current generative models, producing high-quality removal samples proved highly challenging. Ultimately, we selected two optimal regions, ears and hats, and developed tailored prompt generation strategies for their removal tasks. Multimodal language models were leveraged to analyze hair color, length, and style features in target images, enabling dynamic generation of context-matched hair descriptions. \textbf{3) Copy-Move.} Source samples included real images and spliced fake images. Each sample underwent position-shifting of associated regions (only manipulated areas were displaced in fake images, with corresponding masks covering two regions).

\vspace{-1mm}
\subsection{Dataset Statistics}
\label{ssec:dataset statistics}
\vspace{-1mm}

Our dataset comprises three manipulation categories: splicing, removal, and copy-move. The splicing portion is generated with six diffusion models—SD1.5, SD2, SD3, SDXL, FLUX, and Kandinsky 2.1—in two variants (single-region and multi-region), totaling 2040k images split into 1560k for training and 480k for testing. The removal subset contains 13980 images, with 90.8\% ear removals and 9.2\% hat removals. The copy-move subset has 38238 images, evenly divided between two categories.

\begin{figure*}[t]
    \centering
    \includegraphics[width=0.98\linewidth]{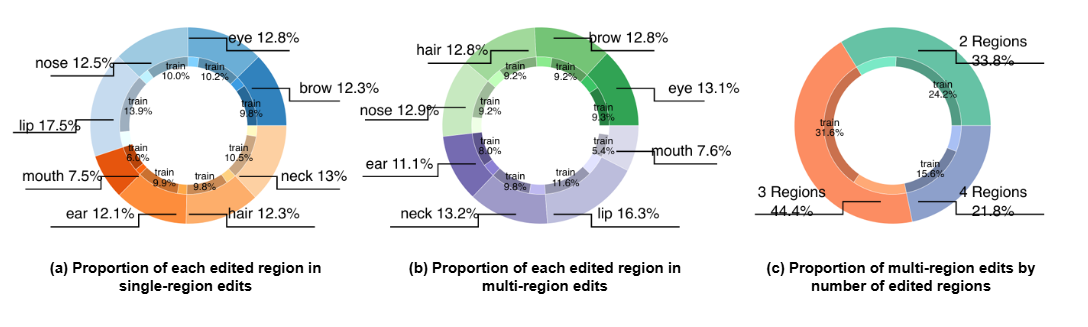}
    \vspace{-15pt}
    \caption{Distribution of the splicing subset. Panel (a) shows the proportion of each edited region in single-region edits; panel (b) shows the proportion of each edited region in multi-region edits; panel (c) shows the distribution of multi-region edits by number of edited regions.}
    \label{fig:dataStatistics}
    \vspace{-10pt}
\end{figure*}

Fig.~\ref{fig:dataStatistics} summarizes distributions for the splicing subset: panel (a) shows per-region shares for single-region edits (approximately uniform across the eight regions); panel (b) shows per-region shares for multi-region edits; and panel (c) shows the distribution of the number of edited regions per image in the multi-region setting. Because regions are selected at random in multi-region editing, both the per-attribute and region-count distributions are nearly uniform overall.

\vspace{-2mm}
\section{Benchmark Settings}
\label{sec:Benchmark Settings}

This section demonstrates the benchmark settings for detection methods and evaluation metrics on DiffFace-Edit.

\noindent
\textbf{Detection Methods.} Our benchmark has implemented 8 locators. The methodologies cover a spectrum that is specifically tailored to detect AI-generated faces from Deepfake Videos, GANs, and DMs.

% Inspired by DeepfakeBench\cite{yan2023deepfakebench}, we classify both detectors into four major categories: \underline{\textit{Naive detectors:}} refer to backbone models that can be directly utilized as the detector for
% binary classification, including CNN-based (i.e., Xception\cite{chollet2017xception}). \underline{\textit{Frequency-based:}} explore the frequency domain for forgery detection (i.e., SPSL\cite{liu2021spatial}, SRM\cite{luo2021generalizing}). \underline{\textit{Edge-based:}} focus on boundary cues by amplifying gradients and high-frequency edges to reveal subtle editing traces (i.e., IML-ViT\cite{ma2023iml}). \underline{\textit{(Vision-language model)–based:}} leverage joint image–text representations and prompt-based reasoning to detect forgeries beyond pixel-level artifacts, improving cross-generator generalization (i.e., UniFD\cite{ojha2023towards}).

\noindent
\textbf{Evaluation Metrics.}
We followed the experimental settings of IMDL-BenCo\cite{ma2024imdl} and adopted the pixel-level F1 score as the primary evaluation metric for assessing the performance of the IMDL model. By default, the pixel-level F1 score for each model is computed using a fixed threshold of 0.5. For the IMDL model, we further report pixel-level IoU as supplementary evaluation metrics.

\noindent
\textbf{Configurations.}
All experiments were conducted on a dedicated server equipped with 8 NVIDIA L40 GPUs.

\section{Results and Analysis}
\label{sec:Results and Analysis}

\subsection{Evasion Analysis in IMDL}
\vspace{-1mm}

The latest comprehensive IMDL evaluation benchmark, IMDL-BenCo\cite{ma2024imdl}, along with other studies, demonstrates that IMDL is strictly defined as images exhibiting partial semantic differences from the original, excluding purely text-generated content, other post-processing modifications, and non-semantic alterations. However, imperceptible perturbations (e.g., semantically ambiguous samples) caused by current multimodal generation techniques whether unavoidable or easily overlooked can severely attack or evade detectors constrained by existing IMDL definitions. Such evasion phenomena may lead IMDL locators to overfit their presumed forged semantic patterns, thereby completely losing their localization capability for semantically ambiguous samples. To comprehensively analyze this issue, we expanded diverse forgery scenarios (e.g., multi-region and single-region forgeries) and constructed correlated forged image sets adhering to IMDL-BenCo’s\cite{ma2024imdl} stringent definition of IMDL samples. For convenience, in the following sections, forgery samples that strictly comply with the IMDL-BenCo standards will be referred to as `\textbf{SC-samples}', semantically ambiguous samples will be referred to as `\textbf{SA-samples}'.

\begin{table*}[ht!]
\caption{
Evaluation of Methods Trained on SC-samples Generated by SD1.5 on SC-samples Generated by Different Diffusion Models.
}
\vspace{-3pt}
\centering
\setlength{\tabcolsep}{10pt}
\renewcommand{\arraystretch}{1.10}
\scriptsize
\begin{tabular}{c|l|c|c|c|c|c|c|c}
\toprule[1.3pt]
% \rowcolor{mygray}
\multirow{2}{*}{Type} & \multirow{2}{*}{Method} & SD1.5(\%)$\uparrow$ & SD2(\%)$\uparrow$ & SD3(\%)$\uparrow$ & SDXL(\%)$\uparrow$ & FLUX(\%)$\uparrow$ & Kan2.1(\%)$\uparrow$ & Avg.(\%)$\uparrow$ \\ \cmidrule(r){3-9}
% \rowcolor{mygray}
 &  & F1 / IoU & F1 / IoU & F1 / IoU & F1 / IoU & F1 / IoU & F1 / IoU & F1 / IoU \\
\midrule
\multirow{4}{*}{Naive}
& Bayar-resnet \cite{ma2024imdl}
& 0.41 / 0.34 & 0.29 / 0.22 & 0.34 / 0.25 & 0.28 / 0.21 & 0.26 / 0.19 & 0.34 / 0.26 & 0.32 / 0.24 \\
& SegFormer-B2 \cite{ma2024imdl}
& 0.22 / 0.17 & 0.11 / 0.08 & 0.14 / 0.11 & 0.10 / 0.07 & 0.10 / 0.07 & 0.14 / 0.10 & 0.13 / 0.10 \\
& SPAN \cite{hu2020span}
& 0.06 / 0.04 & 0.04 / 0.03 & 0.05 / 0.03 & 0.05 / 0.03 & 0.04 / 0.03 & 0.04 / 0.03 & 0.05 / 0.03 \\
& PSCC-Net \cite{liu2022pscc}
& 0.63 / 0.51 & 0.57 / 0.46 & 0.56 / 0.44 & 0.57 / 0.45 & 0.55 / 0.44 & 0.59 / 0.47 & 0.58 / 0.46 \\
\hline
\multirow{3}{*}{Frequency}
& MVSS-Net \cite{dong2022mvss}
& 0.29 / 0.21 & 0.24 / 0.17 & 0.26 / 0.19 & 0.24 / 0.17 & 0.20 / 0.17 & 0.31 / 0.22 & 0.26 / 0.19 \\
& Mesoscopic Insights \cite{zhu2025mesoscopic}
& \underline{0.89} / - & \underline{0.86} / - & \underline{0.87} / - & \underline{0.86} / - & \underline{0.84} / - & \underline{0.88} / - & \underline{0.85} / - \\
& MMFusion \cite{triaridis2023mmfusion}
& 0.83 / \underline{0.82} & 0.83 / \underline{0.82} & 0.83 / \underline{0.82} & 0.82 / \underline{0.81} & 0.82 / \underline{0.81} & 0.82 / \underline{0.81} & 0.82 / \underline{0.82} \\
\hline
\multirow{1}{*}{Edge}
& IML-Vit \cite{ma2023iml}
& \textbf{0.94} / \textbf{0.91} & \textbf{0.94} / \textbf{0.90} & \textbf{0.95} / \textbf{0.92} & \textbf{0.94} / \textbf{0.91} & \textbf{0.94} / \textbf{0.91} & \textbf{0.95} / \textbf{0.92} & \textbf{0.94} / \textbf{0.91} \\
\bottomrule[1.3pt]
\end{tabular}
\label{tab:eva_sc_sc}
\end{table*}

\begin{table*}[ht!]
\caption{
Evaluation of Methods Trained on SC-samples Generated by SD1.5 on SA-samples Generated by Different Diffusion Models.
}
\vspace{-3pt}
\centering
\setlength{\tabcolsep}{10pt}
\renewcommand{\arraystretch}{1.10}
\scriptsize
\begin{tabular}{c|l|c|c|c|c|c|c|c}
\toprule[1.3pt]
% \rowcolor{mygray}
\multirow{2}{*}{Type} & \multirow{2}{*}{Method} & SD1.5(\%)$\uparrow$ & SD2(\%)$\uparrow$ & SD3(\%)$\uparrow$ & SDXL(\%)$\uparrow$ & FLUX(\%)$\uparrow$ & Kan2.1(\%)$\uparrow$ & Avg.(\%)$\uparrow$ \\ \cmidrule(r){3-9}
% \rowcolor{mygray}
 &  & F1 / IoU & F1 / IoU & F1 / IoU & F1 / IoU & F1 / IoU & F1 / IoU & F1 / IoU \\
\midrule
\multirow{4}{*}{Naive}
& Bayar-resnet \cite{ma2024imdl}
& 0.08 / \underline{0.06} & 0.01 / 0.01 & 0.02 / 0.01 & 0.01 / 0.00 & 0.00 / 0.00 & 0.02 / 0.01 & 0.02 / 0.01 \\
& SegFormer-B2 \cite{ma2024imdl}
& 0.02 / 0.01 & 0.00 / 0.00 & 0.01 / 0.01 & 0.00 / 0.00 & 0.00 / 0.00 & 0.00 / 0.00 & 0.01 / 0.00 \\
& SPAN \cite{hu2020span}
& 0.04 / 0.03 & 0.06 / \underline{0.04} & 0.06 / \underline{0.04} & 0.05 / \underline{0.03} & 0.03 / 0.02 & 0.05 / \underline{0.03} & 0.05 / \underline{0.03} \\
& PSCC-Net \cite{liu2022pscc}
& 0.02 / 0.01 & 0.00 / 0.00 & 0.00 / 0.00 & 0.00 / 0.00 & 0.00 / 0.00 & 0.00 / 0.00 & 0.00 / 0.00 \\
\hline
\multirow{3}{*}{Frequency}
& MVSS-Net \cite{dong2022mvss}
& 0.10 / \textbf{0.07} & \underline{0.12} / \textbf{0.08} & \underline{0.12} / \textbf{0.08} & \underline{0.12} / \textbf{0.08} & \textbf{0.10} / \textbf{0.06} & \underline{0.14} / \textbf{0.09} & \underline{0.12} / \textbf{0.08} \\
& Mesoscopic Insights \cite{zhu2025mesoscopic}
& \textbf{0.28} / - & \textbf{0.27} / - & \textbf{0.26} / - & \textbf{0.29} / - & \underline{0.08} / - & \textbf{0.31} / - & \textbf{0.25} / - \\
& MMFusion \cite{triaridis2023mmfusion}
& \underline{0.11} / 0.05 & 0.08 / 0.01 & 0.11 / \underline{0.04} & 0.09 / 0.02 & \underline{0.08} / 0.01 & 0.09 / 0.02 & 0.09 / \underline{0.03} \\
\hline
\multirow{1}{*}{Edge}
& IML-Vit \cite{ma2023iml}
& 0.05 / 0.03 & 0.06 / \underline{0.04} & 0.04 / 0.03 & 0.04 / \underline{0.03} & 0.05 / \underline{0.03} & 0.05 / \underline{0.03} & 0.05 / \underline{0.03} \\
\bottomrule[1.3pt]
\end{tabular}
\label{tab:eva_sc_sa}
\end{table*}

As shown in Tab.~\ref{tab:eva_sc_sc}, we first trained all IMDL methods on single-region SC-samples generated by SD1.5, then evaluated their inherent localization capabilities by testing them on SC-samples produced by six generative models. The results demonstrate that nearly the most methods achieved high IoU and pixel-level F1 scores across SC-sample subsets. However, when replacing the test samples with SA-samples, which contain inherent generative model perturbations, both IoU and F1 dropped significantly, as shown in Tab.~\ref{tab:eva_sc_sa}. For instance, the MMFusion\cite{triaridis2023mmfusion} model trained on SD1.5 achieved an IoU of 0.82 and F1 of 0.83 when tested on SD2-generated SC-samples, but these metrics plummeted to 0.01 (IoU) and 0.08 (F1) on SD2-generated SA-samples. Models like MVSS-Net achieve robustness by capturing forgery artifacts and minimizing reliance on content semantics.

\vspace{-1mm}
\subsection{Counterintuitive Analysis of Removal Samples} 
\vspace{-1mm}

For removal-type samples in generative manipulations, we observe that their detection difficulty significantly exceeds conventional tampering types, exhibiting counterintuitive behavior (i.e., higher visual consistency but higher model misdetection rates). Due to the environmental constraints in generating removal samples (as detailed in Section 2.1), the limited diversity of forged regions in such samples makes the analysis results an unreliable reference. To address this, we conducted a controlled experiment under identical configurations, using only SA-samples with ear manipulations in both the training and test sets.The results in Tab.\ref{tab:splice-remove} show that the average F1 of the IMDL locators on remove samples is lower than that on splicing samples, indicating that remove samples with counterintuitive characteristics may cause misjudgment of the IMDL locators.

Furthermore, we investigate whether the richness of forged regions in training samples impacts detection performance. For comparison, we introduced two additional experimental setups: 1) Samples with 1 randomly selected forged region types. 2) Samples with 5 randomly selected forged region types. The results in Tab.\ref{tab:diff region} demonstrate that as the diversity of tampered regions increases, the model faces greater learning challenges, leading to a decline in localization accuracy.

\begin{table}[tb!]
\centering
\caption{Performance of models trained on ear-splicing fakes (splicing subset) and ear-removal fakes (removal subset), each evaluated on the corresponding test split.}
\label{tab:splice-remove}
\begin{adjustbox}{width=0.75\linewidth}
  \begin{tabular}{l|cc|cc}
  \toprule
  \multirow{2}{*}{Method} & \multicolumn{2}{c}{Splice-Ear}  & \multicolumn{2}{c}{Remove-Ear} \\ \cmidrule(r){2-3}\cmidrule(r){4-5} 
   & F1 & IoU & F1 & IoU \\
  \midrule
  Bayar-ResNet\cite{ma2024imdl} & 0.44 & 0.33 & 0.00 & 0.00\\ 
  SegF-B2\cite{ma2024imdl} & 0.69 & \underline{0.56} & 0.00 & 0.00\\
  SPAN\cite{hu2020span} & 0.02 & 0.01 & 0.02 & 0.01\\
  PSCC-Net\cite{liu2022pscc} & 0.54 & 0.38 & 0.53 & \underline{0.38}\\
  MVSS-Net\cite{dong2022mvss} & 0.43 & 0.33 & 0.45 & 0.33\\
  Mesos.\cite{zhu2025mesoscopic}& \underline{0.70} & - & \underline{0.71} & -\\
  MMFusion\cite{triaridis2023mmfusion}& \textbf{0.78} & \textbf{0.68} & \textbf{0.77} & \textbf{0.67}\\
  IML-ViT\cite{ma2023iml} & 0.44 & 0.33 & 0.52 & 0.36 \\
  \midrule
  \rowcolor{mygray}
  Avg & 0.51 & - & 0.38 & -\\
  \bottomrule
  \end{tabular}
  \end{adjustbox}
\end{table}

\vspace{-1mm}
\subsection{Cross-Manipulation Experiments}
\vspace{-1mm}

\begin{table}[tb!]
\centering
\caption{The performance of IMDL models on samples with different richness levels.}
\label{tab:diff region}
\begin{adjustbox}{width=0.75\linewidth}
  \begin{tabular}{l|cc|cc|cc}
  \toprule
  \multirow{2}{*}{Method} & \multicolumn{2}{c}{One-Region}  & \multicolumn{2}{c}{Five-Region} & \multicolumn{2}{c}{Eight-Region} \\ \cmidrule(r){2-3}\cmidrule(r){4-5}\cmidrule(r){6-7} 
   & F1 & IoU & F1 & IoU & F1 & IoU \\
  \midrule
  Bayar-ResNet\cite{ma2024imdl} & 0.86 & 0.78 & 0.10 & 0.08 & 0.01 & 0.01\\ 
  SegF-B2\cite{ma2024imdl} & 0.83 & 0.77 & 0.07 & 0.06 & 0.04 & 0.03\\
  SPAN\cite{hu2020span} & 0.21 & 0.15 & 0.07 & 0.04 & 0.04 & 0.03\\
  PSCC-Net\cite{liu2022pscc} & 0.87 & 0.79 & \textbf{0.52} & \underline{0.37} & 0.25 & 0.16\\
  MVSS-Net\cite{dong2022mvss} & 0.59 & 0.48 & 0.17 & 0.13 & 0.11 & 0.08\\
  Mesos.\cite{zhu2025mesoscopic}& 0.86 & - & 0.31 & - & \underline{0.29} & -\\
  MMFusion\cite{triaridis2023mmfusion}& \textbf{0.92} & \textbf{0.88} & \underline{0.50} & \textbf{0.45} & \textbf{0.38} & \textbf{0.32}\\
  IML-ViT\cite{ma2023iml} & \underline{0.88} & \underline{0.80} & 0.04 & 0.03 & 0.26 & \underline{0.20}\\
  \midrule
  \rowcolor{mygray}
  Avg & 0.75 & - & 0.22 & - & 0.17 & -\\
  \bottomrule
  \end{tabular}
  \end{adjustbox}
\end{table}

Compared to AIGC detectors, IMDL localizers are more susceptible to both semantic overfitting and positional overfitting (i.e., over-adaptation to specific tampered regions). Results in Tab.\ref{tab:copy-move domain} shows that the edge-enhanced localizer IML-Vit demonstrated the highest localization performance, whereas the naive localizer SPAN failed to effectively localize the forged regions.

\begin{table}[tb!]
\centering
\captionof{table}{Localization performance of IMDL localizers on the Copy-Move Subset.Pristine-CopyMove denotes IMDL-BenCo\cite{ma2024imdl}–compliant samples, whereas Synthesis-CopyMove denotes samples produced by applying copy-move operations to SA-samples.}
\label{tab:copy-move domain}
\begin{adjustbox}{width=0.75\linewidth}
  \begin{tabular}{l|cc|cc}
  \toprule
  \multirow{2}{*}{Method} & \multicolumn{2}{c}{Pristine-CopyMove}  & \multicolumn{2}{c}{Synthesis-CopyMove} \\ \cmidrule(r){2-3}\cmidrule(r){4-5} 
   & F1 & IoU & F1 & IoU \\
  \midrule
  Bayar-ResNet\cite{ma2024imdl} & 0.86 & \underline{0.77} & 0.66 & 0.51\\ 
  SegF-B2\cite{ma2024imdl} & 0.84 & 0.73 & 0.83 & 0.72\\
  SPAN\cite{hu2020span} & 0.01 & 0.01 & 0.02 & 0.01\\
  PSCC-Net\cite{liu2022pscc} & 0.79 & 0.67 & 0.72 & 0.58\\
  MVSS-Net\cite{dong2022mvss} & 0.79 & 0.68 & 0.57 & 0.44\\
  Mesos.\cite{zhu2025mesoscopic}& \underline{0.92} & - & \underline{0.87} & -\\
  MMFusion\cite{triaridis2023mmfusion}& 0.75 & 0.75 & 0.75 & \underline{0.75}\\
  IML-ViT\cite{ma2023iml} & \textbf{0.95} & \textbf{0.91} & \textbf{0.90} & \textbf{0.83} \\
  \midrule
  \rowcolor{mygray}
  Avg & 0.74 & - & 0.67 & -\\
  \bottomrule
  \end{tabular}
  \end{adjustbox}
\end{table}

\vspace{-2mm}
\section{Conclusion}
\label{sec:conclusion}
\vspace{-2mm}

We propose DiffFace-Edit, a large-scale, high-quality facial forgery dataset focused on
partial editing, to address limitations of existing datasets that underestimate the risks associated with facial forgeries. Our dataset comprises over 2M facial images. Each image maintains high semantic consistency with its original counterpart, guided by diverse
prompts. We conduct extensive experiments using DiffFace-Edit
and establish a facial forgery detection benchmark.

% \noindent
% \textbf{Potential Ethical Considerations.} The pristine faces in our
% dataset are sourced from publicly accessible dataset. We have rigorously reviewed all prompts to ensure
% that they do not describe specific biometric details. Each
% generated image has been carefully examined to align with
% societal values. We will try our best to control the acquisition procedure of DiffFace-Edit to mitigate potential misuse.

% Below is an example of how to insert images. Delete the ``\vspace'' line,
% uncomment the preceding line ``\centerline...'' and replace ``imageX.ps''
% with a suitable PostScript file name.
% -------------------------------------------------------------------------

%\vfill
%\pagebreak

\bibliographystyle{IEEEbib}
\small
\bibliography{strings,refs}

\begin{thebibliography}{10}

\bibitem{ding2025fairadapter}
Feng Ding, Jun Zhang, Xinan He, and Jianfeng Xu,
\newblock ``Fairadapter: Detecting ai-generated images with improved fairness,''
\newblock in {\em ICASSP 2025-2025 IEEE International Conference on Acoustics, Speech and Signal Processing (ICASSP)}. IEEE, 2025, pp. 1--5.

\bibitem{pang2024heterogeneous}
Meng Pang, Binghui Wang, Mang Ye, Yiu-Ming Cheung, Yintao Zhou, Wei Huang, and Bihan Wen,
\newblock ``Heterogeneous prototype learning from contaminated faces across domains via disentangling latent factors,''
\newblock {\em IEEE Transactions on Neural Networks and Learning Systems}, vol. 36, no. 4, pp. 7169--7183, 2024.

\bibitem{zhou2025breaking}
Yue Zhou, Xinan He, KaiQing Lin, Bin Fan, Feng Ding, and Bin Li,
\newblock ``Breaking latent prior bias in detectors for generalizable aigc image detection,''
\newblock {\em arXiv preprint arXiv:2506.00874}, 2025.

\bibitem{lin2024preserving}
Li~Lin, Xinan He, Yan Ju, Xin Wang, Feng Ding, and Shu Hu,
\newblock ``Preserving fairness generalization in deepfake detection,''
\newblock in {\em Proceedings of the IEEE/CVF conference on computer vision and pattern recognition}, 2024, pp. 16815--16825.

\bibitem{pang2025unified}
Meng Pang, Wenjun Zhang, Yang Lu, Yiu-ming Cheung, and Nanrun Zhou,
\newblock ``A unified multi-domain face normalization framework for cross-domain prototype learning and heterogeneous face recognition,''
\newblock {\em IEEE Transactions on Information Forensics and Security}, 2025.

\bibitem{ding2021anti}
Feng Ding, Guopu Zhu, Yingcan Li, Xinpeng Zhang, Pradeep~K Atrey, and Siwei Lyu,
\newblock ``Anti-forensics for face swapping videos via adversarial training,''
\newblock {\em IEEE Transactions on Multimedia}, vol. 24, pp. 3429--3441, 2021.

\bibitem{bird2024cifake}
Jordan~J Bird and Ahmad Lotfi,
\newblock ``Cifake: Image classification and explainable identification of ai-generated synthetic images,''
\newblock {\em IEEE Access}, vol. 12, pp. 15642--15650, 2024.

\bibitem{wang2023dire}
Zhendong Wang, Jianmin Bao, Wengang Zhou, Weilun Wang, Hezhen Hu, Hong Chen, and Houqiang Li,
\newblock ``Dire for diffusion-generated image detection,''
\newblock in {\em Proceedings of the IEEE/CVF International Conference on Computer Vision}, 2023, pp. 22445--22455.

\bibitem{zhu2023genimage}
Mingjian Zhu, Hanting Chen, Qiangyu Yan, Xudong Huang, Guanyu Lin, Wei Li, Zhijun Tu, Hailin Hu, Jie Hu, and Yunhe Wang,
\newblock ``Genimage: A million-scale benchmark for detecting ai-generated image,''
\newblock {\em Advances in Neural Information Processing Systems}, vol. 36, pp. 77771--77782, 2023.

\bibitem{cheng2024diffusion}
Harry Cheng, Yangyang Guo, Tianyi Wang, Liqiang Nie, and Mohan Kankanhalli,
\newblock ``Diffusion facial forgery detection,''
\newblock in {\em Proceedings of the 32nd ACM international conference on multimedia}, 2024, pp. 5939--5948.

\bibitem{bhattacharyya2024diffusion}
Chaitali Bhattacharyya, Hanxiao Wang, Feng Zhang, Sungho Kim, and Xiatian Zhu,
\newblock ``Diffusion deepfake,''
\newblock {\em arXiv preprint arXiv:2404.01579}, 2024.

\bibitem{ding2024disrupting}
Feng Ding, Zihan Jiang, Yue Zhou, Jianfeng Xu, and Guopu Zhu,
\newblock ``Disrupting anti-spoofing systems by images of consistent identity,''
\newblock {\em IEEE Signal Processing Letters}, 2024.

\bibitem{fan2025generating}
Bing Fan, Feng Ding, Guopu Zhu, Jiwu Huang, Sam Kwong, Pradeep Atrey, and Siwei Lyu,
\newblock ``Generating higher-quality anti-forensics deepfakes with adversarial sharpening mask,''
\newblock {\em ACM Transactions on Multimedia Computing, Communications and Applications}, vol. 21, no. 6, pp. 1--18, 2025.

\bibitem{he2025vlforgery}
Xinan He, Yue Zhou, Bing Fan, Bin Li, Guopu Zhu, and Feng Ding,
\newblock ``Vlforgery face triad: Detection, localization and attribution via multimodal large language models,''
\newblock {\em arXiv preprint arXiv:2503.06142}, 2025.

\bibitem{ma2024imdl}
Xiaochen Ma, Xuekang Zhu, Lei Su, Bo~Du, Zhuohang Jiang, Bingkui Tong, Zeyu Lei, Xinyu Yang, Chi-Man Pun, Jiancheng Lv, et~al.,
\newblock ``Imdl-benco: A comprehensive benchmark and codebase for image manipulation detection \& localization,''
\newblock {\em Advances in Neural Information Processing Systems}, vol. 37, pp. 134591--134613, 2024.

\bibitem{he2021forgerynet}
Yinan He, Bei Gan, Siyu Chen, Yichun Zhou, Guojun Yin, Luchuan Song, Lu~Sheng, Jing Shao, and Ziwei Liu,
\newblock ``Forgerynet: A versatile benchmark for comprehensive forgery analysis,''
\newblock in {\em Proceedings of the IEEE/CVF conference on computer vision and pattern recognition}, 2021, pp. 4360--4369.

\bibitem{lee2020maskgan}
Cheng-Han Lee, Ziwei Liu, Lingyun Wu, and Ping Luo,
\newblock ``Maskgan: Towards diverse and interactive facial image manipulation,''
\newblock in {\em Proceedings of the IEEE/CVF conference on computer vision and pattern recognition}, 2020, pp. 5549--5558.

\bibitem{hu2020span}
Xuefeng Hu, Zhihan Zhang, Zhenye Jiang, Syomantak Chaudhuri, Zhenheng Yang, and Ram Nevatia,
\newblock ``Span: Spatial pyramid attention network for image manipulation localization,''
\newblock in {\em Computer Vision--ECCV 2020: 16th European Conference, Glasgow, UK, August 23--28, 2020, Proceedings, Part XXI 16}. Springer, 2020, pp. 312--328.

\bibitem{liu2022pscc}
Xiaohong Liu, Yaojie Liu, Jun Chen, and Xiaoming Liu,
\newblock ``Pscc-net: Progressive spatio-channel correlation network for image manipulation detection and localization,''
\newblock {\em IEEE Transactions on Circuits and Systems for Video Technology}, vol. 32, no. 11, pp. 7505--7517, 2022.

\bibitem{dong2022mvss}
Chengbo Dong, Xinru Chen, Ruohan Hu, Juan Cao, and Xirong Li,
\newblock ``Mvss-net: Multi-view multi-scale supervised networks for image manipulation detection,''
\newblock {\em IEEE Transactions on Pattern Analysis and Machine Intelligence}, vol. 45, no. 3, pp. 3539--3553, 2022.

\bibitem{zhu2025mesoscopic}
Xuekang Zhu, Xiaochen Ma, Lei Su, Zhuohang Jiang, Bo~Du, Xiwen Wang, Zeyu Lei, Wentao Feng, Chi-Man Pun, and Ji-Zhe Zhou,
\newblock ``Mesoscopic insights: orchestrating multi-scale \& hybrid architecture for image manipulation localization,''
\newblock in {\em Proceedings of the AAAI Conference on Artificial Intelligence}, 2025, vol.~39, pp. 11022--11030.

\bibitem{triaridis2023mmfusion}
Kostas Triaridis, Konstantinos Tsigos, and Vasileios Mezaris,
\newblock ``Mmfusion: Combining image forensic filters for visual manipulation detection and localization,''
\newblock {\em arXiv preprint arXiv:2312.01790}, 2023.

\bibitem{ma2023iml}
Xiaochen Ma, Bo~Du, Zhuohang Jiang, Ahmed Y~Al Hammadi, and Jizhe Zhou,
\newblock ``Iml-vit: Benchmarking image manipulation localization by vision transformer,''
\newblock {\em arXiv preprint arXiv:2307.14863}, 2023.

\end{thebibliography}

\end{document}